\pdfoutput=1 

\documentclass[11pt,a4paper]{article}
\usepackage[hyperref]{acl2020}
\usepackage{xurl}  
\usepackage{booktabs}  
\usepackage{tabularx}
\usepackage{graphicx}
\graphicspath{ {./images/} }
\usepackage{enumitem}
\usepackage{times}
\usepackage{latexsym}
\usepackage{amsmath}
\usepackage{multirow}
\usepackage{soul}
\usepackage{color}
\usepackage[capitalise,noabbrev]{cleveref}
\crefrangelabelformat{section}{#3#1#4-#5\crefstripprefix{#1}{#2}#6}

\usepackage{microtype}

\aclfinalcopy 

\definecolor{ExampleColor}{RGB}{255, 200, 200}
\definecolor{ResourceColor}{RGB}{200, 255, 200}
\definecolor{PromptColor}{RGB}{200, 200, 255}
\definecolor{RelaxationColor}{RGB}{224, 224, 224}
\definecolor{IndexColor}{RGB}{255, 200, 255}
\newcommand{\hlc}[2][yellow]{{\sethlcolor{#1}\hl{#2}}}

\title{The Turking Test: Can Language Models Understand Instructions?}

\author{Avia Efrat \\
  Tel-Aviv University \\
  \texttt{avia.efrat@gmail.com} \\\And
  Omer Levy \\
  Tel-Aviv University \\
  Facebook AI Research \\
  \texttt{omerlevy@gmail.com} \\}

\date{}

\begin{document}
\maketitle

\begin{abstract}
Supervised machine learning provides the learner with a set of input-output examples of the target task.
Humans, however, can also learn to perform new tasks from \textit{instructions} in natural language.
Can machines learn to understand instructions as well?
We present the Turking Test, which examines a model's ability to follow natural language instructions of varying complexity.
These range from simple tasks, like retrieving the $n$th word of a sentence, to ones that require creativity, such as generating examples for SNLI and SQuAD in place of human intelligence workers (``turkers'').
Despite our lenient evaluation methodology, we observe that a large pretrained language model performs poorly across all tasks.
Analyzing the model's error patterns reveals that the model tends to ignore explicit instructions and often generates outputs that cannot be construed as an attempt to solve the task.
While it is not yet clear whether instruction understanding can be captured by traditional language models, the sheer expressivity of instruction understanding makes it an appealing alternative to the rising few-shot inference paradigm.

\end{abstract}

\section{Introduction}

One of the fundamental problems in AI is how to build a model that can generalize to new, previously unseen \textit{tasks}.
Recent work \citep{brown2020language} proposes a few-shot inference approach, in which a massive language model, GPT-3, is conditioned on a few input-output examples of a new task, followed by the input we wish the model to process.
This approach works surprisingly well on a wide range of tasks.
While humans can often understand a task from a handful of examples, they can also benefit from a natural language description of \textit{how} to perform it, i.e. \textit{instructions}.
Instructions can describe a task \textit{accurately} and \textit{succinctly}; e.g. ``Write the fourth word of the following sentence'' is a more compact description of a task than a dozen pairs of sentences and their fourth words.
If language models can perform new tasks by conditioning on input-output pairs, can they do so by conditioning on \textit{instructions} as well? 

We propose the Turking Test, a series of instruction-following benchmarks of varying complexity.
We begin with turking tasks (\cref{sec:TurkingTasks}), where the model is tasked with creating valid dataset examples of popular NLP datasets (SNLI, SQuAD, and NewsQA), simulating a task commonly done by laypeople on crowdsourcing platforms.
In \cref{sec:ListingNouns}, we instruct the model to list all the nouns of a given sentence that satisfy a simple condition.
Finally, we ask the model to write the $n$th word/character of a given sentence (\cref{sec:NthElement}).
Throughout the process, we take a lenient evaluation approach to improve the model's chance of being successful, such as selecting the best generation algorithm and instruction set post-hoc.

We observe poor performance across all experiments when applying the Turking Test to GPT-2 \cite{radford2019language}, a 1.5B parameter language model.\footnote{Despite submitting an API request in June 2020, we have not yet received access to GPT-3.}
For example, the model achieves only 2\% accuracy on the simple task of writing the $n$th word, which can be easily executed by an elementary school child.
We also observe that the model largely ignores explicit restrictions and conditions that appear in the instructions, achieving slightly higher accuracy on open-ended tasks than those with well-defined answers.

We conclude by looking at instruction understanding as a learning paradigm, and show that it is a strict generalization of the few-shot paradigm. While few-shot examples can be embedded into instructions, one clear advantage of the instruction paradigm is the ability to provide an explicit signal of what \textit{not} to do, where examples can only hope to provide this negative signal implicitly. 
While it is not yet clear whether instruction understanding can be captured by traditional language models,
achieving such an ability will greatly extend the range of tasks that we can readily handle and alleviate the need to annotate large task-specific datasets.

\section{Instruction Understanding}
\label{sec:Method}

In an \textit{instruction understanding task} (IUT), a model is provided with an input $I_{x}$ that describes in natural language a desired output $o$. $I_{x}$ is comprised of a template $I$ that we call \textit{instructions}, which is instantiated with a \textit{resource} $x$ to form $I_{x}$. $I$ may or may not contain input-output examples.

A simple IUT could be ``Write the fourth word of the sentence $x$.'' In this case the instructions are the entire string except $x$ (hence $I$ is a template), and the input $I_{x}$ can be formed by setting the resource $x$ to be ``Today was a good day''.

\begin{figure}[t]
  \centering
  \includegraphics[width=\columnwidth]{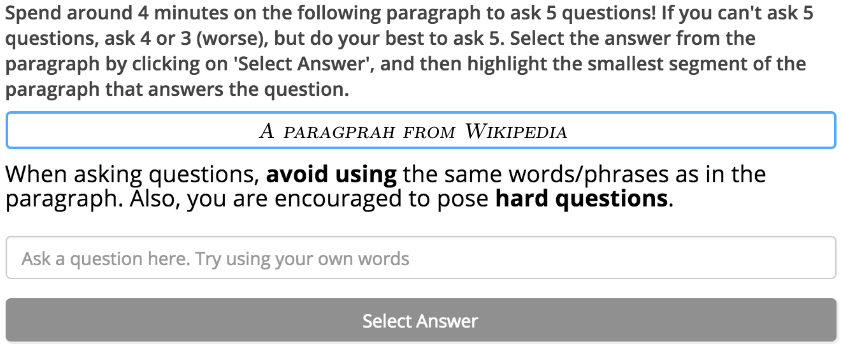}
  \caption{The original instructions of SQuAD 1.1, on which we base our SQuAD IUT (see \cref{sec:TurkingTasks}).}
  \label{fig:OriginalSquadTask}
\end{figure}

A more challenging IUT could be annotating examples for a typical NLP dataset, a task often carried out by human intelligence workers (``turkers'') on crowdsourcing platforms such as Amazon Mechanical Turk. In this case, $I$ are the instructions given to the turker, and $x$ is the specific text used to create the dataset example (e.g. a paragraph from Wikipedia in the case of SQuAD). \cref{fig:OriginalSquadTask} illustrates such a task. These IUTs may have multiple different outputs $o$ that are all equally valid.

\subsection{Experiments and Tasks}

We perform three sets of experiments to assess instruction understanding. Each experiment consists of two or three tasks of a similar nature.

\paragraph{Turking Tasks (\cref{sec:TurkingTasks})}
Test the model's ability to annotate popular datasets such as SQuAD \cite{DBLP:journals/corr/RajpurkarZLL16} and SNLI \cite{snli:emnlp2015}, basing the instructions on the original annotation guidelines and using the development set for resources.

\paragraph{Listing Nouns (\cref{sec:ListingNouns})}
Focuses on much shorter instructions: listing all the nouns of a given sentence that satisfy a simple condition.

\paragraph{Retrieving an Element by Index (\cref{sec:NthElement})}
Consists of the simplest tasks, where the model is instructed to output the $n$th word or character of a sentence.

\subsection{Authoring the Instructions} \label{subsec:AuthoringInstructions}

To increase the model's chances, we manually refine each task's instructions using a small subset of resources, and use the version on which the model performed best. 
All the instructions in our experiments go through this iterative refinement process, regardless of whether a task has a preexisting version of instructions (like the turking tasks) or not.

\subsection{Model} \label{subsec:model}

We use GPT-2 \citep{radford2019language} as the language model in our experiments via Hugging Face's \citep{wolf2019huggingfaces} implementation.
At the time of writing, we have yet to be granted access to GPT-3 \citep{brown2020language}.
We produce outputs with both greedy decoding and nucleus sampling \citep{DBLP:journals/corr/abs-1904-09751} with $p=0.9$, generating up to a maximum number of tokens (tuned per task) or until the model predicts EOS.

\subsection{Evaluation} \label{subsec:evaluation}

We use a group of 10 college-graduate English-speaking annotators to evaluate each of the model's outputs and determine whether it indeed follows the given instructions.

We introduce three relaxations to increase the model's chance of success:

\paragraph{The Prefix Rule:} If a prefix of an output is itself a valid output, then the entire output is considered valid. We use this rule since the model rarely generates EOS, so even if it manages to generate a valid output, it is often followed by irrelevant text (see \cref{fig:FullTurkingTask}).

\paragraph{The Introduction Rule:}
The model is allowed to generate an ``introduction'' before the actual answer, such as ``Here are the nouns of the sentence:'' (see \cref{fig:FullTurkingTask}).

\paragraph{Oracle Generation Algorithm:} 
We report the model's performance on each task using the generation algorithm (greedy or nucleus sampling with $p=0.9$) that leads to the best result; the best algorithm is not necessarily the same across all tasks.

\begin{figure}
\centering
\small
\begin{tabular}{@{}p{\columnwidth}@{}}
\toprule
Write questions about the highlights of a story. \\
\\
Steps \\
1. Read the highlights \\
2. Write questions about the highlights \\
\\
\hlc[ExampleColor]{Example \\
\\
Highlights \\
$\bullet$ Sarah Palin from Alaska meets with McCain \\
$\bullet$ Fareed Zakaria says John McCain did not put country first with his choice \\
$\bullet$ Zakaria: This is ``hell of a time'' for Palin to start thinking about national, global issues} \\
\\
Questions \\
The questions can refer directly to the highlights, for example: \\
\hlc[ExampleColor]{$\bullet$ Where is Palin from? \\
$\bullet$ What did Fareed say about John McCain's choice? \\
$\bullet$ Who is thinking about global issues?} \\
\\
Questions must always be related to the highlights but their answers don't have to be in the highlights. You can assume that the highlights summarize a document which can answer other questions for example: \\
\hlc[ExampleColor]{$\bullet$ What was the meeting about? \\
$\bullet$ What was McCain's choice? \\
$\bullet$ What issues is Palin thinking about?} \\
\\
Other Rules \\
$\bullet$ Do not re-use the same or very similar questions. \\
$\bullet$ Questions should be written to have short answers. \\
$\bullet$ Do not write ``how'' nor ``why'' type questions since their answers are not short. ``How far/long/many/much'' are okay. \\
 \\
\hlc[PromptColor]{Here are the highlights:} \\
\hlc[ResourceColor]{$\bullet$ Math geeks and others celebrate Pi Day every March 14 \\
$\bullet$ Pi, or roughly 3.14, is the ratio of circumference to diameter of a circle \\
$\bullet$ The Pi Day holiday idea started at the Exploratorium museum in San Francisco \\
$\bullet$ Albert Einstein was also born on March 14} \\
 \\
\hlc[PromptColor]{Write questions about them:} \\
\midrule
\hlc[RelaxationColor]{Here are questions about the highlights:} \\
1. When is Pi Day celebrated? \\
2. What is the value of Pi up to the second decimal digit? \\
\hlc[RelaxationColor]{Another thing is that Pi is important in mathematics.} \\
\bottomrule
\end{tabular}

\caption{
An example of the NewsQA IUT with a potential output.
\textbf{Input:}
The input $I_{x}$ is composed of the instruction template $I$, which is fixed, and the variable resource $x$ (green).
Note that $I$ contains an example resource and corresponding valid outputs (red), which are different from the actual resource $x$.
The instructions $I$ are modeled after the original instructions of NewsQA \cite{DBLP:journals/corr/TrischlerWYHSBS16}, but replace visual HTML cues (e.g ``Ask a question here.'' from \cref{fig:OriginalSquadTask}) with purely textual prompts (blue), as without them the model's accuracy was 0\% on every turking task.
\textbf{Output:}
The first line (gray) demonstrates the introduction rule, and the last line (gray) demonstrates the prefix rule. The middle two lines are the only ones being evaluated.
}
\label{fig:FullTurkingTask}
\end{figure}

\section{Turking Tasks}\label{sec:TurkingTasks}

Turking tasks are tasks that are normally carried out by human intelligence workers (``turkers'') on crowdsourcing platforms such as Amazon Mechanical Turk. Their instructions are given in natural language, and they require basic language skills that the vast majority of adults posses.

\subsection{Tasks} \label{subsec:TurkingTasks}

Our first experiment consists of three turking tasks.
We replicate the annotation processes of SNLI \citep{snli:emnlp2015}, SQuAD \citep{DBLP:journals/corr/RajpurkarZLL16}, and NewsQA \citep{DBLP:journals/corr/TrischlerWYHSBS16} -- each time replacing the human turker with the language model.
These popular datasets have high inter-annotator agreement, indicating that laypeople can understand their instructions reasonably well.
While there is prior work on learning to ask reading comprehension questions from thousands of examples \cite{du-etal-2017-learning}, our goal here is to test how well a model performs when given the same information as human annotators.

\paragraph{SNLI}
Given a caption of a photo (without the photo itself), produce three sentences (hypotheses): one that is \textit{definitely} a true description of the photo (entailment), one that \textit{might} be a true description of the photo (neutral), and one that is definitely \textit{not} a true description of the photo (contradiction).

\paragraph{SQuAD}
Given a paragraph from Wikipedia, write question-answer pairs, where the questions are about the paragraph and their answers are spans of text from the paragraph.

\paragraph{NewsQA}
Given highlights from a news article (without the article itself), write questions about either the highlights or the article (the turker is to assume what is reasonable to be written in the article given the highlights). \cref{fig:FullTurkingTask} shows our NewsQA IUT.




\paragraph{Evaluation Criteria}
In addition to the prefix and the introduction rules (\cref{subsec:evaluation}), we further relax the requirements of a valid output.
For SNLI, the generated hypotheses do not need to be explicitly labeled entailment/neutral/contradiction; as long as the first three sentences can form a valid SNLI example (in any order), the output is valid.
For SQuAD, while the original instructions encourage writing five question-answer pairs, we require only three questions (and \textit{no answers}), matching the requirements of the other question annotation task, NewsQA.\footnote{\cref{apx:instructions} contains the instructions for all our IUTs.}

\subsection{Experimental Setting} \label{subsec:TurkingExperimentalSetting}

We generate 500 outputs for each task. The outputs are generated using 100 different resources with five nucleus ($p=0.9$) generations per resource, as preliminary experiments show that greedy generation leads to 0\% accuracy across all turking tasks. Each annotator was given 50 outputs to score. We found very high inter-annotator agreement (98\% observed agreement) based on a sample from all three tasks.

For each task we only select resources that the original human turkers found easy to annotate.
In SNLI, we use captions (premises) that have complete agreement (5/5) on each of their hypotheses.
In SQuAD, we use paragraphs that have at least five questions.
In NewsQA, we use highlights that have at least six \textit{answerable} questions.



\subsection{Results} \label{subsec:TurkingResults}

\begin{table}[t]
    \small
    \centering
    \begin{tabular}{@{}lrr@{}}
        \toprule
        \textbf{Task} & \textbf{Accuracy} & \textbf{Consistency} \\
        \midrule
        SNLI  & 0.4\% & 12.0\% \\
        SQuAD & 0.2\% & 6.0\% \\
        NewsQA & 4.2\% & 22.0\% \\
        \bottomrule
    \end{tabular}
    \caption{Results on the turking tasks. An output is \textit{accurate} if it correctly follows the instructions of the IUT. An output is \textit{form consistent} if it produces an output in the correct \textit{format}, but does not necessarily have correct \textit{content} (see \cref{subsec:TurkingResults} for a formal definition).}
    \label{tab:TurkingResults}
\end{table}

\cref{tab:TurkingResults} shows the results on the turking tasks.
Results on SNLI and SQuAD are very poor, with only 0.4\% and 0.2\% of the outputs correctly following the instructions. Results on NewsQA (4.2\%), albeit very low as well, are still more than an order of magnitude higher than the other question generation IUT, SQuAD. 

One possible explanation is that NewsQA contains an example in the instruction template (one-shot), whereas SQuAD does not (zero-shot).
However, SNLI's instructions also contain an example, and its results are on par with SQuAD.
An alternative explanation is that SQuAD's instructions require the questions to be answerable by a \textit{span} from the resource, while NewsQA does not.
This means that a NewsQA output can be valid just by asking reasonable on-topic questions.
We suspect that the open-ended nature of the NewsQA IUT is particularly favorable to language models, as subsequent experiments show that the model fails to perform very simple IUTs with well-defined outputs. (\cref{sec:ListingNouns,sec:NthElement}).

\paragraph{Form Consistency}
As turking tasks are rather challenging IUTs, we further evaluate performance by introducing a more lenient metric, \textit{form consistency}.
Formally, an output $o$ is \textit{form consistent} with respect to an input $I_{x}$ if there could exist a resource $x^{*}$ in the context of the task such that $o$ would be considered accurate with respect to $I_{x^{*}}$.
\cref{fig:FormConsistency} shows an example of a form consistent (but incorrect) SNLI output.

\begin{figure}[t]
  \centering
  \small
  \begin{tabular}{@{}ll@{}}
    \toprule
    Resource $x$  & Five children are playing a video game. \\
    \midrule
     & A man walking. \\
    Output $o$  & A man wearing a blue hat is walking. \\
     & A man sitting on a bench. \\
    \midrule
    Imaginary $x^*$ & A man wearing a hat is walking. \\
    \bottomrule
  \end{tabular}
  \caption{An example of a form consistent output of the SNLI turking task. Although the output $o$ is not correct w.r.t the given resource $x$, there still could exist a resource $x^{*}$, e.g. ``a man wearing a hat is walking'', that $o$ is correct with respect to it. With this $x^{*}$, the generated sentences are (in order) entailed, neutral, and contradicting.} 
  \label{fig:FormConsistency}
\end{figure}

\cref{tab:TurkingResults} shows the model's form consistency on each task, based on a random sample of 50 examples.
Although form consistency sets a low bar for the model, the vast majority of its outputs are not even in the right format.
For example, we observe that 94\% of SQuAD's outputs are not even form consistent, i.e. do not contain three questions that could be answered by spans from any reasonable Wikipedia paragraph.
Even for the less restrictive NewsQA IUT, we find that 78\% of the outputs cannot be tied to any reasonable news article.

\subsection{Error Analysis} \label{subsec:TurkingErrorAnalysis}

\begin{figure*}[t]
\centering
\small
\begin{tabularx}{\textwidth}{@{}XX@{}}
\toprule

\textbf{(a) Continuing the Resource (NewsQA)} & \\
\midrule
\textit{Highlights} \newline
$\bullet$ Anatomy Lab app allows users to dissect a virtual cadaver \newline 
$\bullet$ Researchers say app is aimed at medical and anatomy students \newline
$\bullet$ The app has so far sold more than 3,000 copies &
\textit{Questions} \newline
$\bullet$ Scientists say that the new apps offer a new way of studying human anatomy \newline
$\bullet$ Students can also use the apps to answer more traditional lab questions \\
\midrule

\textbf{(b) Continuing the Resource (SNLI)} & \\
\midrule
\textit{Premise} \newline
A group of people are gathered in the woods around trees and a ladder. &
\textit{Hypotheses} \newline
On the left is a tree. On the right is a ladder. The person is sitting at the top. A rainstorm is rolling in. Around him are two groups of people having a picnic. \\
\midrule

\textbf{(c) Continuing the Instructions (SNLI)} & \\
\midrule
\textit{Premise} \newline
A young man plays a racing video game. &
\textit{Hypotheses} \newline
$\bullet$ See if your caption reveals what the \hlc[ExampleColor]{dog} is doing. Example: If the man and \hlc[ExampleColor]{dog} are playing a racing video game, then they might be playing a billiards game. \\
\midrule

\textbf{(d) Non Resource-Specific Questions (SQuAD)} & \\
\midrule
\textit{Paragraph} \newline 
There are over 10,000 objects made from silver or gold in the collection, the display (about 15\% of the collection) is divided into secular and sacred covering both... &
\textit{Questions} \newline
1. Is the passage interesting/rich/interesting? \newline
2. Was the passage organized or extensive? \newline
3. Was the passage topical or deliberately obscure? \\
\midrule

\textbf{(e) Unrelated Questions (NewsQA)} & \\
\midrule
\textit{Highlights} \newline
$\bullet$ Among items auctioned Saturday: jacket, song lyrics, sketches, dental mold \newline
$\bullet$ Famous rhinestone-studded glove fetched \$420,000... &
\textit{Questions} \newline
$\bullet$ Does this item fit with \hlc[ExampleColor]{Palin's} personality? \newline
$\bullet$ Do they embody her political sensibilities? \newline
$\bullet$ Are they a match with \hlc[ExampleColor]{McCain's} image of family values? \\
\midrule

\textbf{(f) Grounding on Instruction Example (SNLI)} & \\
\midrule
\textit{Premise} \newline
A woman wearing sunglasses is holding a pink cellphone to her ear. &
\textit{Hypotheses} \newline
A woman wearing sunglasses is holding a pink cellphone to her ear. \newline
\hlc[ExampleColor]{Some puppies are running to catch a stick.} \newline
\hlc[ExampleColor]{There are animals outdoors.} \\
\midrule

\textbf{(g) Ignoring Instructions (NewsQA)} & \\
\midrule
\textit{Highlights} \newline
$\bullet$ Inter Milan go two points clear of second-placed Roma with 3-1 win over lowly Atalanta \newline
$\bullet$ Wesley Sneijder in doubt for Wednesday's trip... &
\textit{Questions} \newline
$\bullet$ Why are the European champions so close to relegation? \newline
$\bullet$ Why are Napoli's 100,000 fans not up in arms? \newline
$\bullet$ Why does Roma sit fourth? \\
\bottomrule

\end{tabularx}
\caption{Common mistake patterns on the turking tasks. The resource $x$ is on the left and output $o$ is on the right. Text in \textit{italics} is not a part of the resource or the output.
}
\label{fig:TurkingErrorAnalysis}
\end{figure*}

\begin{figure*}[t]
    \centering
    \small
    \begin{tabular}{cccp{0.42\textwidth}l}
        \toprule
        \textbf{Case} & \textbf{Sound} & \textbf{Number} & \textbf{Instructions} & \textbf{Possible Output} \\
        \midrule
        \multirow{2}{*}{\textit{lower}} & & \multirow{2}{*}{\textit{singular}} & List all the singular nouns from the sentence ``We have a meeting on Sunday'' that start with a lowercase letter. & \multirow{2}{*}{The nouns are: meeting} \\
        \midrule
        \multirow{2}{*}{\textit{upper}} & \multirow{2}{*}{\textit{consonant}} & & List all the nouns from the sentence ``We have a meeting on Sunday'' that start with an uppercase consonant. & \multirow{2}{*}{We, Sunday} \\
        \midrule
         & \multirow{2}{*}{\textit{vowel}} & \multirow{2}{*}{\textit{plural}} & List all the plural nouns from the sentence ``We have a meeting on Sunday'' that start with a vowel. &  \\
        \bottomrule
    \end{tabular}
    \caption{Three examples of the conditioned noun listing task with possible valid outputs for the resource ``We have a meeting on Sunday.'' No nouns satisfy the third example's condition, and it is therefore omitted (see \cref{subsec:ListNounsExperimentalSetting}).}
    \label{tab:ListConditionedNounsExamples}
\end{figure*}

We randomly sample 150 outputs from each turking task and analyze common errors.
\cref{fig:TurkingErrorAnalysis} shows recurring mistake patterns.

\paragraph{Continuing the Resource}
Instead of following the instructions, the model just generates a continuation of the resource, such as adding more highlights in NewsQA (see \cref{fig:TurkingErrorAnalysis}a).
This is more prevalent when the resource is short, occurring in 24\% and 17\% of NewsQA's and SNLI's outputs respectively, but only 3\% of the time in SQuAD.

\paragraph{Continuing the Instructions}
The model generates more instructions instead of following them (see \cref{fig:TurkingErrorAnalysis}c).
This is particularly prevalent in SNLI, where it occurs in 19\% of all outputs.

\paragraph{Ungrounded Questions}
Whereas the previous two patterns are much less common in SQuAD, in 38\% of its outputs \textit{all} the questions are either so general that they could apply to any resource (23\%, see \cref{fig:TurkingErrorAnalysis}d) or unrelated to the resource (15\%).
This also happens in NewsQA 9\% of the time (\cref{fig:TurkingErrorAnalysis}e).

\paragraph{Grounding on the Instruction Example}
Both SNLI's and NewsQA's instructions contain an example resource-output pair (see \cref{fig:FullTurkingTask} for NewsQA). 11\% of SNLI's and 6\% of NewsQA's outputs are grounded on these examples, and they are always inaccurate. The grounding is either explicit (copying the example verbatim, highlighted in \cref{fig:TurkingErrorAnalysis}f) or implicit (highlighted in \ref{fig:TurkingErrorAnalysis}c and \ref{fig:TurkingErrorAnalysis}e).

\paragraph{Ignoring Explicit Restrictions}
NewsQA's instructions explicitly forbid using ``why'' and some ``how'' questions (see \cref{fig:FullTurkingTask}). Nevertheless, 49\% of NewsQA's outputs contain at least one such question, indicating the model does not follow superficial restrictions in the instructions (see \cref{fig:TurkingErrorAnalysis}g).

\section{Listing Nouns} \label{sec:ListingNouns}

Following the language model's poor performance on turking tasks (\cref{sec:TurkingTasks}), we test whether it can follow the instructions of a simpler linguistic task: listing nouns of a given sentence.

\subsection{Tasks} \label{subsec:ListingNounTasks}

The first noun listing task is straightforward:
\begin{align*}
\text{List all the nouns of the sentence ``$x$''.}
\end{align*}
The second task is a variation on the first: given a sentence, list all nouns \textit{that satisfy a condition}. A condition is a conjunction of two properties from the following three: \textbf{case} (starts with an uppercase/lowercase letter), \textbf{sound} (starts with a consonant/vowel), and \textbf{number} (singular/plural). There are 12 possible conditions, three of which are illustrated in \cref{tab:ListConditionedNounsExamples}.


\subsection{Experimental Setting} \label{subsec:ListNounsExperimentalSetting}

We sample 100 sentences from GLUE's diagnostic dataset \citep{DBLP:journals/corr/abs-1804-07461}, which spans five different domains and diverse linguistic phenomena, and use them as resources to instantiate inputs for both tasks.
We pair every resource to each of the 12 possible conditions and filter cases where no nouns satisfy the condition, leaving a total of 100 examples for unconditioned noun listing and 689 examples in the conditioned setting.

For each input, we perform one greedy generation and one $p=0.9$ nucleus generation, limiting the output to $|x| + 20$ tokens (where $|x|$ is the number of tokens in the resource sentence).
As nucleus generation led to 0\% accuracy on both tasks, we only present the results of the greedy generation (see Oracle Generation Algorithm in \cref{subsec:evaluation}).

\subsection{Results}

\begin{figure}[t]
  \centering
  \includegraphics[width=\columnwidth]{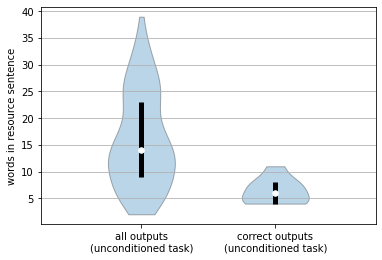}
  \caption{The distribution of sentence length (in number of words) over all resources (100 sentences) versus those resources for which the model correctly predicted the full list of nouns in the unconditioned task (9 sentences). Medians in white, middle quartiles in black.}
  \label{fig:LengthToSuccessLnLnp}
\end{figure}

\cref{tab:ListNounsResults} shows the results on the noun listing tasks.
Although the instructions of these tasks are shorter and simpler than in turking tasks, results are still poor, with only 9\% accuracy on the unconditioned task, and with successes highly skewed towards shorter sentences, as shown in \cref{fig:LengthToSuccessLnLnp}.
Moreover, adding a simple condition to the task's instructions results in a substantial drop in performance, as only 3.5\% of the conditioned task's outputs are correct, with no significant performance difference across different conditions.
This resonates the findings from \cref{subsec:TurkingErrorAnalysis}, as again the model struggles adhering to simple restrictions in the instructions.

\begin{table}[t]
\small
\centering
\begin{tabular}{@{}lr@{}}
  \toprule
   \textbf{Task}   &   \textbf{Accuracy} \\
  \midrule
   List all nouns   &  9.0\% \\
   List all nouns satisfying a condition    &  3.5\% \\
  \bottomrule
\end{tabular}
\caption{Performance on noun listing tasks using greedy decoding.}
\label{tab:ListNounsResults}
\end{table}

\subsection{Error Analysis} \label{subsec:ListingNounAnalysis}

In both tasks greedy generation outputs are sometimes comprised of repeated copies of either the resource or the entire instruction. It occurs in 10\% of the unconditioned task's outputs and in 30\% of the conditioned task's outputs, always resulting in an invalid output. While it has been shown that nucleus sampling can mitigate repetitions \citep{DBLP:journals/corr/abs-1904-09751}, we find it leads to 0\% accuracy on both tasks.

\section{Retrieving Elements by Index} \label{sec:NthElement}

Listing nouns (\cref{sec:ListingNouns}) still requires some linguistic knowledge. For our last experiment, we use instructions so simple that they could be implemented by a single line of code in Python: retrieving the $n$th word or character in a sentence.

\subsection{Tasks}

We present the model with two retrieval tasks:
\vspace{6pt}

\noindent
(1) Write the $n$th word from the sentence ``$x$''.
\vspace{3pt}

\noindent
(2) Write the $n$th character from the sentence ``$x$''.
\vspace{6pt}

\noindent
While optimizing the instructions (see \cref{subsec:AuthoringInstructions}), we found that better results were achieved with a slight variation of the second task:
\vspace{6pt}

\noindent
Write down the $n$th letter from the sentence ``$x$'':

\subsection{Experimental Setting} \label{subsec:NthElementExperimentalSetting}

We use all 832 sentences from GLUE's diagnostic dataset (\citealt{DBLP:journals/corr/abs-1804-07461}) as resources to instantiate inputs for both tasks. We limit the word index to 20 and the character index to 40. For the character task, we never ask for an index of a whitespace character. For each sentence we perform one greedy generation and one $p=0.9$ nucleus generation, for a total of 1664 outputs per task. We generate $|x| + 20$ tokens for each input, where $|x|$ is the number of tokens in the resource sentence. Since the evaluation of these tasks is simple and unambiguous, we preform it ourselves. As in \cref{sec:ListingNouns}, we only report results based on greedy generation since nucleus sampling was significantly worse on both tasks (see Oracle Generation Algorithm in \cref{subsec:evaluation}).

\subsection{Results}

\begin{table}[t]
\small
\centering
\begin{tabular}{@{}lrr@{}}
  \toprule
   \textbf{Task}    &   \textbf{Accuracy} & \textbf{Baseline} \\
  \midrule
  $n$th word       &  2.0\% & 6.4\% \\
  $n$th character  &  1.3\% & 13.8\% \\
  \bottomrule
\end{tabular}
\caption{Performance on the $n$th element tasks using greedy decoding. The baseline writes the most common word (``the'') or character (``e'') in the language.}
\label{tab:NthElementResults}
\end{table}

\cref{tab:NthElementResults} shows the results on the $n$th element tasks. The model does not exceed 2\% accuracy on either task, well below the simple baseline of writing the most common word or character.

\subsection{Error Analysis}

In \cref{subsec:ListingNounAnalysis} we saw that in the conditioned noun listing task either the resource or the entire input are repeated in the output 30\% of the time.
Here the situation is even worse: repetition occurs in 46\% and 59\% of the $n$th word and $n$th character outputs respectively, always resulting in an incorrect output.
The fact that such a large percentage of outputs is comprised of senseless repetitions indicates that the model fails to understand these trivial instructions.

To analyze common repetition patterns, we sample 50 repetition outputs from each task. Even though these tasks are similar and have almost identical instructions, we find that their repetition patterns significantly differ, suggesting the model is hyper-sensitive to small changes in the instructions. For example, while 96\% of the sampled $n$th character outputs begin with the resource, \textit{none} of the sampled $n$th word outputs do. In addition, while 54\% of the sampled $n$th character outputs are repetitions of \textit{only} the resource, 100\% of the sampled $n$th word outputs involve repetitions of \textit{both} the resource and some variant of the instructions.
\section{Discussion: The Instruction Paradigm}

Currently, the prevailing paradigm in NLP is supervised learning from many labeled examples, often supported by pretraining à la BERT \citep{DBLP:journals/corr/abs-1810-04805}.
Recently, GPT-3 \citep{brown2020language} demonstrated the effectiveness of an alternative paradigm, few-shot inference, by conditioning a massive pretrained language model with only a handful of task-specific examples at inference time.
In this work, we explore a third alternative, \textit{the instruction paradigm}, where the model is conditioned on a natural language description of the task.

The instruction paradigm is a strict generalization of GPT-3's few-shot paradigm.
Instructions \textit{generalize} few-shot inference because they may contain examples.
In fact, \cref{sec:TurkingTasks} shows two tasks (SNLI and NewsQA), where examples of expected inputs and outputs are embedded into the instructions.
This generalization is \textit{strict} because instructions can convey information that examples cannot.
For example, instructions can explicitly state what is \textit{not} a valid output (e.g. ``Do not write `why' questions...'' -- NewsQA), whereas examples, even more than a few, can only hope to provide this negative signal implicitly.

While the instruction paradigm is very expressive and natural, it also poses a serious challenge.
Throughout this paper, we observe that a large language model fails to follow a series of gradually simpler instructions.
While it is possible that a larger language model or one trained on instructional language may perform better, it might also be the case, as posited by \citet{bender-koller-2020-climbing}, that models may require some form of grounding beyond the linguistic form to learn how to follow instructions and pass the Turking Test.

    








\section*{Acknowledgements}

This work was supported in part by the Blavatnik Fund, the Alon Scholarship, and the Tel Aviv University Data Science Center.
We are grateful for the help of Uri Shaham, Yuval Kirstain, Avia Amitay, Zohar Arnon, Dafna Barzilai, Sivan Barzily, Pavel Brodsky, Anna Landa, Tomer Landsberger, Jonathan Rosen, and Ran Ziv in human evaluation and feedback.

\bibliography{anthology,acl2020}
\bibliographystyle{acl_natbib}

\newpage
\appendix

\section{All Instructions} \label{apx:instructions}

This appendix contains the full version for all the instructions we used throughout \cref{sec:TurkingTasks,sec:ListingNouns,sec:NthElement}. All instructions are additionally appended with the newline.

\begin{figure*}[htb]
\centering
\small
\frenchspacing  
\begin{tabularx}{\textwidth}{>{\ttfamily}X}
\toprule
We will show you the caption for a photo. We will not show you the photo. Using only the caption and what you know about the world: \newline \newline
• Write one alternate caption that is definitely a true description of the photo. Example: For the caption "Two dogs are running through a field." you could write "There are animals outdoors." \newline
\newline
• Write one alternate caption that might be a true description of the photo. Example: For the caption "Two dogs are running through a field." you could write "Some puppies are running to catch a stick." \newline \newline
• Write one alternate caption that is definitely a false description of the photo. Example: For the caption "Two dogs are running through a field. you could write "The pets are sitting on a couch." This is different from the maybe correct category because it's impossible for the dogs to be both running and sitting. \newline \newline \newline
Here's the caption: "\textit{\hlc[ResourceColor]{resource}}". \newline
According to the above instructions, write three alternate captions below, one that is definitely a true description of the photo, one that might be a true description of the photo, and one that is definitely a false description of the photo: \\
\bottomrule
\end{tabularx}

\caption{
Full instructions of the SNLI IUT (\cref{sec:TurkingTasks}). To instantiate an input, the highlighted part in italics is replaced with a resource -- a caption from the SNLI development set.}
\label{fig:SNLIFullInstructions}
\end{figure*}

\begin{figure*}[htb]
\centering
\small
\frenchspacing  
\begin{tabularx}{\textwidth}{>{\ttfamily}X}
\toprule
Read the following paragraph and ask 5 questions about it! If you can't ask 5 questions, ask 4 or 3 (worse), but do your best to ask 5. The answers to these questions must be answerable by a span of text from the passage. \newline \newline
\textit{\hlc[ResourceColor]{resource}} \newline \newline
When asking questions, avoid using the same words/phrases as in the paragraph. Also, you are encouraged to pose hard questions. \newline
Now according to the instructions above, ask five questions about the paragraph: \\
\bottomrule
\end{tabularx}

\caption{
Full instructions of the SQuAD IUT (\cref{sec:TurkingTasks}). To instantiate an input, the highlighted part in italics is replaced with a resource -- a Wikipedia paragraph from the SQuAD development set.}
\label{fig:SQuADFullInstructions}
\end{figure*}

\begin{figure*}[htb]
\centering
\small
\frenchspacing  
\begin{tabularx}{\textwidth}{>{\ttfamily}X}
\toprule
Write Questions From A Summary \newline \newline
Overview \newline \newline
Write questions about the highlights of a story. \newline \newline
Steps \newline \newline
1. Read the highlights \newline
2. Write questions about the highlights \newline \newline
Example \newline \newline
Highlights \newline
• Sarah Palin from Alaska meets with McCain \newline
• Fareed Zakaria says John McCain did not put country first with his choice \newline
• Zakaria: This is "hell of a time" for Palin to start thinking about national, global issues \newline \newline
Questions \newline
The questions can refer directly to the highlights, for example: \newline \newline
• Where is Palin from? \newline
• What did Fareed say about John McCain's choice? \newline
• Who is thinking about global issues? \newline \newline
Questions must always be related to the highlights but their answers don't have to be in the highlights. You can assume that the highlights summarize a document which can answer other questions for example: \newline \newline
• What was the meeting about? \newline
• What was McCain's choice? \newline
• What issues is Palin thinking about? \newline \newline
Other Rules \newline \newline
• Do not re-use the same or very similar questions. \newline
• Questions should be written to have short answers. \\
\hspace{1.8em} • Do not write "how" nor "why" type questions since their answers are not short. "How far/long/many/much" are okay. \newline \newline \newline
Here are the highlights: \newline
\textit{\hlc[ResourceColor]{resource}} \newline
Write questions about them: \\
\bottomrule
\end{tabularx}

\caption{
Full instructions of the NewsQA IUT (\cref{sec:TurkingTasks}). To instantiate an input, the highlighted part in italics is replaced with a resource -- article highlights from the NewsQA development set.}
\label{fig:NewsQAFullInstructions}
\end{figure*}

\begin{figure*}[htb]
\centering
\small
\frenchspacing  
\begin{tabularx}{\textwidth}{>{\ttfamily}X}
\toprule
List all the nouns of the sentence "\textit{\hlc[ResourceColor]{resource}}". \\
\bottomrule
\end{tabularx}

\caption{
Full instructions of the unconditioned noun listing IUT (\cref{sec:ListingNouns}). To instantiate an input, the highlighted part in italics is replaced with a resource -- a sentence from the GLUE diagnostic dataset.}
\label{fig:ListNounsUnconditionedFullInstructions}
\end{figure*}

\begin{figure*}[htb]
\centering
\small
\frenchspacing  
\begin{tabularx}{\textwidth}{>{\ttfamily}X}
\toprule

List all the nouns from the sentence "\textit{\hlc[ResourceColor]{resource}}" that start with a lowercase consonant. \\
\midrule
List all the nouns from the sentence "\textit{\hlc[ResourceColor]{resource}}" that start with an uppercase consonant. \\
\midrule
List all the nouns from the sentence "\textit{\hlc[ResourceColor]{resource}}" that start with a lowercase vowel. \\
\midrule
List all the nouns from the sentence "\textit{\hlc[ResourceColor]{resource}}" that start with an uppercase vowel. \\
\midrule
List all the singular nouns from the sentence "\textit{\hlc[ResourceColor]{resource}}" that start with a lowercase letter. \\
\midrule
List all the plural nouns from the sentence "\textit{\hlc[ResourceColor]{resource}}" that start with a lowercase letter. \\
\midrule
List all the singular nouns from the sentence "\textit{\hlc[ResourceColor]{resource}}" that start with an uppercase letter. \\
\midrule
List all the plural nouns from the sentence "\textit{\hlc[ResourceColor]{resource}}" that start with an uppercase letter. \\
\midrule
List all the singular nouns from the sentence "\textit{\hlc[ResourceColor]{resource}}" that start with a consonant. \\
\midrule
List all the plural nouns from the sentence "\textit{\hlc[ResourceColor]{resource}}" that start with a consonant. \\
\midrule
List all the singular nouns from the sentence "\textit{\hlc[ResourceColor]{resource}}" that start with a vowel. \\
\midrule
List all the plural nouns from the sentence "\textit{\hlc[ResourceColor]{resource}}" that start with a vowel. \\
\bottomrule
\end{tabularx}

\caption{
Full instructions of the conditioned noun listing IUT (\cref{sec:ListingNouns}), for each of the 12 possible conditions. To instantiate an input, the highlighted part in italics is replaced with a resource -- a sentence from the GLUE diagnostic dataset.}
\label{fig:ListNounsConditionedFullInstructions}
\end{figure*}

\begin{figure*}[htb]
\centering
\small
\frenchspacing  
\begin{tabularx}{\textwidth}{>{\ttfamily}X}
\toprule
"Write the \textit{\hlc[IndexColor]{n}}th word from the sentence "\textit{\hlc[ResourceColor]{resource}}". \\
\bottomrule
\end{tabularx}

\caption{
Full instructions of the \textit{n}th word IUT (\cref{sec:NthElement}). To instantiate an input, \textit{\hlc[IndexColor]{n}} is replaced with a number between 1 an 20 (up to the sentence's length in words) and \textit{\hlc[ResourceColor]{resource}} is replaced with a sentence from the GLUE diagnostic dataset.}
\label{fig:NthWordFullInstructions}
\end{figure*}

\begin{figure*}[htb]
\centering
\small
\frenchspacing  
\begin{tabularx}{\textwidth}{>{\ttfamily}X}
\toprule
"Write down the \textit{\hlc[IndexColor]{n}}th letter from the sentence "\textit{\hlc[ResourceColor]{resource}}". \\
\bottomrule
\end{tabularx}

\caption{
Full instructions of the \textit{n}th character IUT (\cref{sec:NthElement}). To instantiate an input, \textit{\hlc[IndexColor]{n}} is replaced with a number between 1 an 40 (up to the sentence's length in characters) and \textit{\hlc[ResourceColor]{resource}} is replaced with a sentence from the GLUE diagnostic dataset.}
\label{fig:NthCharacterFullInstructions}
\end{figure*}

\end{document}